\documentclass[a4paper]{llncs}
\usepackage{graphicx}
\usepackage{times}
\usepackage{helvet}
\usepackage{courier}
\usepackage{amsmath, amssymb, mathtools}
\usepackage{color}
\usepackage{tabu}
\usepackage{ulem}
\newtheorem{result}{Result}
\newtheorem{prop}{Proposition}

\usepackage{multirow}
\begin{document}
\title{Improving Generalization of Deep Networks for Inverse Reconstruction of Image Sequences}
%
%
%
%
\vspace{-.3cm}
\author{Sandesh Ghimire\inst{1} \and Prashnna Kumar Gyawali\inst{1}
\and Jwala Dhamala\inst{1} \and John L Sapp\inst{2}\and Milan Horacek\inst{2}\and Linwei Wang\inst{1}
}

\authorrunning{S.Ghimire et. al.} 
\institute{ Rochester Institute of Technology, Rochester, NY 14623, USA \\
\email{sg9872@rit.edu},
\texttt{www.sandeshgh.com}
 \and Dalhousie University, Halifax, NS, Canada\\
}
\maketitle              
\vspace{-.6cm}
\begin{abstract}
Deep learning networks have shown state-of-the-art performance in many image reconstruction problems. However, it is not well understood what properties of representation and learning may improve the generalization ability of the network. 
In this paper, 
we propose that the generalization ability of an encoder-decoder network for inverse reconstruction 
can be improved in two means. 
First, 
drawing from analytical learning theory, 
we theoretically show that a stochastic latent space will improve the ability of a network to generalize to test data outside the training distribution. 
Second, 
following the information bottleneck principle, 
we show that a latent representation minimally informative of the input data will help a network generalize to unseen input variations 
that are irrelevant to the output reconstruction. 
Therefore, we present a sequence image reconstruction network optimized by a variational approximation of the information bottleneck principle with stochastic latent space. 
In the application setting of 
reconstructing the sequence of cardiac transmembrane potential from body-surface potential, 
we assess the two types of generalization abilities of the presented network 
against its deterministic counterpart. 
The results demonstrate that the generalization ability of an inverse reconstruction network can be improved by 
stochasticity as well as the information bottleneck.

\end{abstract}
\vspace{-.3cm}
\section{Introduction}
{There has been an upsurge of deep learning approaches for} 
{traditional image reconstruction problems}
in computer vision and medical imaging \cite{lucas18}. {Examples include image} denoising \cite{mao16}, inpainting \cite{pathak2016context}, 
and {medical} image reconstructions {across} a variety of  modalities such as magnetic resonance imaging  
\cite{zhu18}
and computed tomography
\cite{han2016deep}.
Despite {state-of-the-art performances brought by}
{these deep} neural networks, 
{their ability to reconstruct from data not seen in the training distribution} 
 is not well understood. To date, very limited work has investigated the generalization ability of these image reconstruction networks from a theoretical perspective, or provided insights into what aspects of representation and learning may improve the ability of these networks to  
generalize outside the training data.

In this paper, we {take an information theoretic perspective -- along with analytical learning theory -- to} {investigate and} improve the generalization ability of {deep} image reconstruction {networks}. 
Let $\boldsymbol{x}$ be the original image 
and $\boldsymbol{y}$ be the measurement 
obtained from $\boldsymbol{x}$ by some transformation process. To  reconstruct $\boldsymbol{x}$ from $\boldsymbol{y}$, {we adopt a common deep {encoder-decoder} architecture 
\cite{pathak2016context,zhu18} where a latent representation $\boldsymbol{w}$ is first inferred from $\boldsymbol{y}$ before being used for the reconstruction of $\boldsymbol{x}$.  Our objective is to learn transformations that are general, possibly learning the underlying generative process rather than focusing on every detail in training examples. To this end, we propose that the generalization ability of a deep reconstruction network can be improved from two means: 
1) the ability to generalize to data $\boldsymbol{y}$ that are 
generated from $\boldsymbol{x}$ (and thereby $\boldsymbol{w}$) outside the training distribution;} 
and 2) the ability to generalize to unseen variations in data $\boldsymbol{y}$ 
that are introduced during the measurement process but irrelevant to $\boldsymbol{x}$.  

For the first type of generalization ability, 
we hypothesize that it can be improved by using stochastic {instead of deterministic} latent representations. We support this hypothesis by {the} analytical learning theory \cite{kawaguchi18analytical}, showing that stochastic latent space helps to learn a decoder that is less sensitive to perturbations in the latent space {and thereby} leads to better generalization. 
For the second type of generalization ability, 
we hypothesize that it can be improved if the encoder compresses the input measurement into a minimal latent representation (\textit{codes} in information theory), containing {only} the necessary information for $\boldsymbol{x}$ to be reconstructed. 
{To obtain a minimal representation from $\boldsymbol{y}$ that is maximally informative of $\boldsymbol{x}$,}
we adopt the information bottleneck theory formulated in \cite{tishby2000} to 
maximize {the} mutual information between {the} latent code {$\boldsymbol{w}$} and $\boldsymbol{x}$,  $I(\boldsymbol{x},\boldsymbol{w})$, {while putting a} constraint on the mutual information between $\boldsymbol{y}$ and $\boldsymbol{w}$
such that $I(\boldsymbol{w},\boldsymbol{y})<I_0$. This can be achieved by minimizing the following {objective}:
\begin{align}
    \label{ll}
    loss_{IB} = -I(\boldsymbol{x;w})+ \beta I(\boldsymbol{w;y})
\end{align}
where $\beta$ is the Lagrange multiplier. 
Based on these two primary hypotheses, we present a  deep image reconstruction network optimized by a variational approximation of the information bottleneck principle with stochastic latent space. 

While the presented network applies for general reconstruction problems, 
we test it on the sequence  reconstruction of cardiac transmembrane potential (TMP)  
from 
{high-density body-surface electrocardiograms (ECGs)} 
\cite{wang10}.
Given the sequential nature of the problem, 
 we use long short-term memory (LSTM) networks in both the encoder and decoder,  {with} two {alternative} architectures to compress the temporal information into vector latent space. 
We tackle two specific challenges 
regarding the generalization of the reconstruction. 
First, because the problem 
is ill-posed, it has been important to 
constrain the reconstruction with prior physiological knowledge of 
TMP dynamics \cite{greensite98,wang10,ghimire2018generative}.  
This however made it difficult 
to generalize to  physiological conditions outside those specified by the prior knowledge. By using the stochastic latent space, we demonstrate the ability of the presented method to generalize outside the 
physiological knowledge provided in the training data.  
Second, because 
the generation of ECGs depends on  heart-torso geometry, 
it has been difficult for 
existing methods to generalize beyond a patient-specific setting. 
By the use of the information bottleneck principle, we demonstrate the robustness of the presented network to geometrical variations in ECG data and therefore a unique ability to generalize to unseen subjects.  
These generalization abilities are tested in two controlled synthetic datasets as well as
a real-data feasibility study.  {We hope that} these findings may 
initiate {more theoretical and systematic} investigations of the generalization ability of {deep networks in} {image reconstruction problems}.
\vspace{-.1cm}
\section{Related Work}
\vspace{-.2cm}
Deep neural networks have become popular in medical image reconstructions across different  modalities such as computed tomography 
\cite{han2016deep},
magnetic resonance imaging  \cite{zhu18}, 
and ultrasound 
\cite{luchies2018deep}. 
{Some of these inverse reconstruction networks are 
based on an encoder-decoder structure \cite{han2016deep,zhu18}, similar to that 
investigated in this paper. 
Among these, 
 {the presented}
 work is the closest to Automap \cite{zhu18} {in that} the output image is reconstructed directly from the {input} measurements without {any} intermediate domain-specific   transformations. 
 However, these {existing} works {have not investigated either the use of} stochastic architectures {or the} information bottleneck principle
 to improve {the ability of the network to generalize outside the training distributions}.


The presented theoretical analysis of
stochasticity in generalization utilizes analytical learning theory \cite{kawaguchi18analytical}, which is fundamentally different from classic statistical learning theory in that it is strongly instance-dependent. While statistical learning theory deals with data-independent generalization bounds or data-dependent bounds for certain hypothesis space, analytical learning theory 
provides the bound on how well a model learned from a dataset should perform on true (unknown) measures of variable of interest. This makes it aptly suitable for measuring the generalization ability of stochastic latent space for the given problem and data.

The presented variational formulation of {the information bottleneck principle} is closely related to {that presented in} \cite{alemi2017}. However, our work differs in {three} primary aspects. First, 
we {investigate image reconstruction tasks in which} the role of information bottleneck has not been clearly understood. Second, 
we define  generalization ability in two different categories, and  
provide theoretical as well as empirical evidence on how  stochastic latent space can improve the network's generalization ability in a way different from the information bottleneck. Finally, we extend the setting of static image classification 
to} 
image sequences, {in which the latent representation needs to be compressed from temporal information within} the whole sequence. 

{To learn temporal relationship in ECG/TMP sequences, we consider two sequence encoder-decoder architectures.}
One is 
commonly used in  
language translation \cite{sutskever14}, where the code from {the last unit of the last LSTM encoder layer} is used as the latent vector representation to reconstruct $\boldsymbol{x}$.
We also present a second architecture where fully connected layers are used to {compress all the hidden codes of the last LSTM layer into a} latent vector representation. {This is in concept similar to the 
attention mechanism}
\cite{bahdanau14}
to selectively use  information 
from all the hidden LSTM codes for decoding. We experimentally compare the generalization ability of using  stochastic versus  deterministic latent vectors in both architectures, which has not been studied before. 

In the application area of cardiac TMP reconstruction, most related to this paper are works constraining the reconstruction with prior temporal knowledge 
in the form of 
physics-based simulation models of TMP \cite{wang10} and, more recently, 
generative models learned from physics-based TMP simulation \cite{ghimire2018generative}. This however to our knowledge is the first work that investigated the use of deep learning for the direct inference of TMP from ECG. This method will also have the unique potential to generalize outside the patient-specific settings and outside pathological conditions included in the prior knowledge.  
\vspace{-.5cm}
\section{Methodology}
\vspace{-.2cm}
{Body-surface electrical potential is produced by TMP in the heart. Their} mathematical relation 
is defined by the quasi-static approximation of electromagnetic theory 
{\cite{plonsey1969bioelectric} and, when solved on patient-specific heart-torso geometry, can be derived as:}  
$\boldsymbol{y}(t)=\boldsymbol{Hx}(t)$, 
where $\boldsymbol{y}(t)$ denotes the {time-varying} body-surface potential map, $\boldsymbol{x}(t)$ the {time-varying} TMP map over the 3D heart muscle, 
and $\boldsymbol{H}$ the measurement matrix specific to the heart-torso geometry of a subject \cite{wang10}. 
{The inverse reconstruction of $\boldsymbol{x}$ from $\boldsymbol{y}$ 
at each time instant is ill-posed, and 
a popular approach is to reconstruct TMP time sequence constrained by prior physiological knowledge of its dynamics  
\cite{ghimire2018generative,wang10,greensite98}. This is the setting considered in this study, in which the deep network learns to reconstruct with prior knowledge 
from pairs of $\boldsymbol{x}(t)$ and $\boldsymbol{y}(t)$ generated by physics-based simulation. Note that it is not possible to obtain real TMP data for training, which further highlights the importance of the network to generalize.  
In what follows,  
we use $\boldsymbol{x}$ and $\boldsymbol{y}$ to represent  
sequence matrices with each column 
denoting the potential map at one time instant. 
Given the joint distribution of TMP and ECG given by $p(\boldsymbol{x},\boldsymbol{y})$,  
the encoder gives us a conditional distribution $p(\boldsymbol{w}|\boldsymbol{y})$. These together defines a joint distribution of $(\boldsymbol{x},\boldsymbol{y}$,$\boldsymbol{w}$):
\begin{align} \vspace{-.1cm}
    p(\boldsymbol{x},\boldsymbol{y},\boldsymbol{w})=p(\boldsymbol{x})p(\boldsymbol{y}|\boldsymbol{x})p(\boldsymbol{w}|\boldsymbol{x},\boldsymbol{y})=p(\boldsymbol{x},\boldsymbol{y})p(\boldsymbol{w}|\boldsymbol{y})
\end{align}
The first term in $loss_{IB}$ in eq.(\ref{ll}) is given by
\begin{align}
\nonumber
    I(\boldsymbol{x};\boldsymbol{w})=\int  p(\boldsymbol{x},\boldsymbol{w})\log (\frac{p(\boldsymbol{x}|\boldsymbol{w})}{p(\boldsymbol{x})})d\boldsymbol{x}d\boldsymbol{w}=H(\boldsymbol{x})+ \int  p(\boldsymbol{x},\boldsymbol{w})\log ({p(\boldsymbol{x}|\boldsymbol{w})})d\boldsymbol{x}d\boldsymbol{w}
\end{align}
where 
$p(\boldsymbol{x}|\boldsymbol{w})=\int \frac{p(\boldsymbol{x},\boldsymbol{w},\boldsymbol{y})}{p(\boldsymbol{w})}d\boldsymbol{y}=\int \frac{p(\boldsymbol{x},\boldsymbol{y})p(\boldsymbol{w}|\boldsymbol{y})}{p(\boldsymbol{w})}d\boldsymbol{y}$ is intractable.  
Letting $q(\boldsymbol{x}|\boldsymbol{w})$ {to} be the variational approximation of {$p(\boldsymbol{x}|\boldsymbol{w})$}, we have:
\begin{align}
\nonumber
\int  p(\boldsymbol{x},&\boldsymbol{w}){\log ({p(\boldsymbol{x}|\boldsymbol{w})})d\boldsymbol{x}d\boldsymbol{w}} =\int { p (\boldsymbol{w})[ p(\boldsymbol{x}|\boldsymbol{w}) \log \frac{p(\boldsymbol{x}|\boldsymbol{w})}{q(\boldsymbol{x}|\boldsymbol{w})}+p(\boldsymbol{x}|\boldsymbol{w})\log q(\boldsymbol{x}|\boldsymbol{w})]d\boldsymbol{x}d\boldsymbol{w}}\\
&=\int p(\boldsymbol{w}) D_{KL}(p(\boldsymbol{x}|\boldsymbol{w})||q(\boldsymbol{x}|\boldsymbol{w})) d\boldsymbol{w} 
+\int p(\boldsymbol{x},\boldsymbol{w})\log q(\boldsymbol{x}|\boldsymbol{w})d\boldsymbol{x}d\boldsymbol{w}
\end{align}
where 
the KL divergence 
in the first term is
non-negative. 
This gives us: 
\begin{align}
    \label{first_term}
    I(\boldsymbol{x};\boldsymbol{w})
    \geq \int p(\boldsymbol{x},\boldsymbol{y},\boldsymbol{w})\log q(\boldsymbol{x}|\boldsymbol{w})]d\boldsymbol{x}d\boldsymbol{y}d\boldsymbol{w}
    =E_{p(\boldsymbol{x},\boldsymbol{y})} [E_{p(\boldsymbol{w}|\boldsymbol{y})} [\log q(\boldsymbol{x}|\boldsymbol{w})]] 
\end{align}
{The} second term in $loss_{IB}$ in {eq.(\ref{ll})} is given by
\begin{align}
\label{KL_encoder}
\nonumber
    I(\boldsymbol{y};\boldsymbol{w})&=\int  p(\boldsymbol{y},\boldsymbol{w})\log (\frac{p(\boldsymbol{w}|\boldsymbol{y})}{p(\boldsymbol{w})})d\boldsymbol{y}d\boldsymbol{w}
    =\int  p(\boldsymbol{y},\boldsymbol{w})\log [\frac{p(\boldsymbol{w}|\boldsymbol{y})r(\boldsymbol{w})}{r(\boldsymbol{w})p(\boldsymbol{w})}]d\boldsymbol{y}d\boldsymbol{w}\\
    &=\int  p(\boldsymbol{y})p(\boldsymbol{w}|\boldsymbol{y})\log (\frac{p(\boldsymbol{w}|\boldsymbol{y})}{r(\boldsymbol{w})})d\boldsymbol{y}d\boldsymbol{w}-D_{KL}(p(\boldsymbol{w})||r(\boldsymbol{w}))\\
    \label{second_term}
    &\leq \int  p(\boldsymbol{y})p(\boldsymbol{w}|\boldsymbol{y})\log (\frac{p(\boldsymbol{w}|\boldsymbol{y})}{r(\boldsymbol{w})})d\boldsymbol{y}d\boldsymbol{w} = E_{p(\boldsymbol{y})}[D_{KL}(p(\boldsymbol{w}|\boldsymbol{y})||r(\boldsymbol{w}))]
\end{align}
Combining eq.(\ref{first_term}) and eq.(\ref{second_term}), we have 
\begin{align}
\label{upper_bound}
loss_{IB} \leq  E_{p(\boldsymbol{x},\boldsymbol{y})} [- E_{p(\boldsymbol{w}|\boldsymbol{y})} \big[\log q(\boldsymbol{x}|\boldsymbol{w})]+\beta D_{KL}(p(\boldsymbol{w}|\boldsymbol{y})||r(\boldsymbol{w}))\big] 
= \mathcal{L}_{IB}
\end{align}
which gives us $\mathcal{L}_{IB}$ to be minimized as an upper bound  of the information bottleneck objective $loss_{IB}$ formulated in eq.(\ref{ll}).

\vspace{-.4cm}
\subsubsection{Parameterization with neural network:}
We model both { $p(\boldsymbol{w}|\boldsymbol{y})$ and $q(\boldsymbol{x}|\boldsymbol{w})$} 
as Gaussian distributions, 
with mean and variance 
parameterized by neural networks: 
\begin{align}
\label{encoder-decoder}
      p_{\boldsymbol{\theta}_1}(\boldsymbol{w}|\boldsymbol{y})=\mathcal{N}(\boldsymbol{w}|\boldsymbol{t}_{\theta_1}(y),\boldsymbol{\sigma_t}^2(\boldsymbol{y}))
      \hspace{1cm}
  q_{\boldsymbol{\theta}_2}(\boldsymbol{x}|\boldsymbol{w})=\mathcal{N}(\boldsymbol{x}|\boldsymbol{g}_{\theta_2}(\boldsymbol{w}),\boldsymbol{\sigma_x}^2(\boldsymbol{w}))
\end{align}
where $\boldsymbol{\sigma_x}^2$ denotes a matrix that consists of the variance of each corresponding element in matrix $\boldsymbol{x}$. This is based on the implicit assumption that each elements in $\boldsymbol{x}$ is independent and Gaussian, 
and similarly for $\boldsymbol{w}$. This gives us: 
\begin{align}
\nonumber
   \mathcal{L}_{IB}(\boldsymbol{\theta}) =  E_{p(\boldsymbol{x},\boldsymbol{y})} [- E_{p_{\boldsymbol{\theta_1}}(\boldsymbol{w}|\boldsymbol{y})} [\log q_{\boldsymbol{\theta}_2}(\boldsymbol{x}|\boldsymbol{w})]+\beta. D_{KL}(p_{\boldsymbol{\theta}_1}(\boldsymbol{w}|\boldsymbol{y})||r(\boldsymbol{w}))] 
\end{align}
where $\boldsymbol{\theta}=\{\boldsymbol{\theta}_1, \boldsymbol{\theta}_2 \}$. 
We use reparameterization $\boldsymbol{w}=\boldsymbol{t}+\boldsymbol{\sigma}_t\odot \boldsymbol{\epsilon}$ as described in \cite{kingma13} to compute 
the inner expectation in the first term. 
The KL divergence in the second term 
is analytically available for two Gaussian distributions. 
We obtain: 
\begin{align}
\nonumber
\mathcal{L}_{IB}(\boldsymbol{\theta}) &= E_{p(\boldsymbol{x},\boldsymbol{y})} \Big[ E_{\boldsymbol{\epsilon}\sim\mathcal{N}(\boldsymbol{0},\boldsymbol{I})}\Big( \sum_i\frac{1}{\boldsymbol{\sigma_x}_{i}^2}(x_i-g_i(\boldsymbol{t}+\boldsymbol{\sigma_t}\odot \boldsymbol{\epsilon}))^2+\log \boldsymbol{\sigma_x}_{i}^2\Big)\\
\label{loss}
&
+\beta . D_{KL}(p_{\boldsymbol{\theta}_1}(\boldsymbol{w}|\boldsymbol{y})||\mathcal{N}(\boldsymbol{w}|\boldsymbol{0},\boldsymbol{I})) \Big]
\end{align}
where $g_i$ is the $i^{th}$ function mapping latent variable to the  $i^{th}$ element of mean of $\boldsymbol{x}$, such that $\boldsymbol{g}_{\boldsymbol{\theta}_2}=[g_1,g_2...g_U]$. The deep network is trained to minimize $\mathcal{L}_{IB}(\boldsymbol{\theta})$ in eq.(\ref{loss}) with respect to network parameters $\boldsymbol{\theta}$.

\begin{figure*}[t!]
\centering
\includegraphics[width=\linewidth]{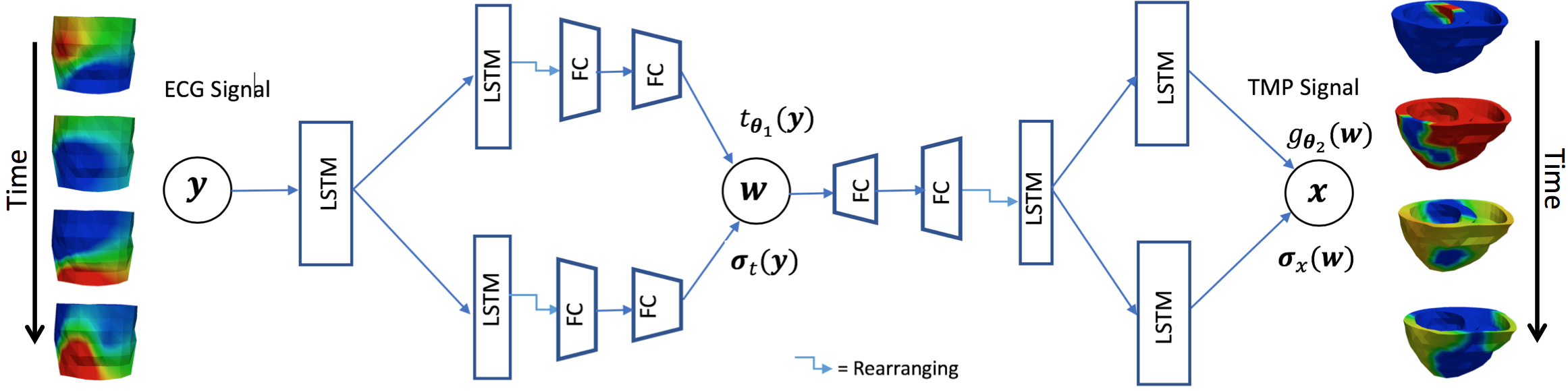}
\caption{\small{Illustration of the presented \textit{svs stochastic} architecture, where both the encoder and the decoder consists of mean and variance networks.}}
\label{architecture}
\end{figure*}
\vspace{-.3cm}
\subsubsection{Network architectures:} 
The sequence reconstruction network is realized using 
long short-term memory (LSTM) neural networks in both the encoder and decoder.
{To compress the time sequence into}
a latent vector  representation,    
we experiment with two {alternative} architectures. 
First, based on {the commonly-used} sequence-to-sequence language translation {model} \cite{sutskever14},{we consider a \textit{svs-L}} architecture {that employs} the hidden code {of the last  unit in the last encoding LSTM layer} as the latent vector representation 
for reconstructing TMP sequences. Second,  we propose {a \textit{svs}} architecture where two fully connected layers are used to {compress all the hidden codes of the last LSTM layer into a} vector representation. {In the decoder, this latent representation is expanded by} two fully-connected layers {before being fed into} LSTM layers as shown in Fig. \ref{architecture}.  

\vspace{-.3cm}
\section{Encoder-Decoder Learning from the Perspective of Analytical Learning Theory}
\vspace{-.2cm}
In this section we look at the encoder-decoder inverse reconstructions using 
analytical learning theory \cite{kawaguchi18analytical}. We start with a general framework and then show that having a stochastic latent space with regularization helps in generalization.

Let $\boldsymbol{z}=(\boldsymbol{y},\boldsymbol{x})$ be an input-output pair, and let $D_n=\{\boldsymbol{z}^{(1)},\boldsymbol{z}^{(2)},...,\boldsymbol{z}^{(n)}\}$ denote the total set of training and validation data where $Z_m \subset D_n$ be the validation set. During training, a neural network learns the parameter $\boldsymbol{\theta}$ by using an algorithm $\mathcal{A}$ and dataset $D_n$, at the end of which we have a mapping $h_{\mathcal{A}(D_n)}(.)$ from $\boldsymbol{y}$ to $\boldsymbol{x}$.
Typically, we stop training when the model performs well in the validation set. To evaluate this performance, we define a prediction error function, $\ell(\boldsymbol{x},h_{\mathcal{A}(D_n)}(\boldsymbol{y}))$ based on our notion of the goodness of prediction. The average validation error is given by ${E_{Z_m}\ell(\boldsymbol{x},h_{\mathcal{A}(D_n)}(\boldsymbol{y}))}$. 
However, there exists a so-called generalization gap between how well the model performs in the validation set versus in the true distribution of the input-output pair. To be precise, let $(\mathcal{Z},\mathcal{S},\mu)$ be a measure space with $\mu$ being a measure on $(\mathcal{Z},\mathcal{S})$. Here, $\mathcal{Z}=\mathcal{Y}\times\mathcal{X}$ denotes the input-output space of all the observations and inverse solutions.  The generalization gap is given by
$
\Delta_g={E_{\mu}\ell(\boldsymbol{x},h_{\mathcal{A}(D_n)}(\boldsymbol{y}))}-{E_{Z_m}\ell(\boldsymbol{x},h_{\mathcal{A}(D_n)}(\boldsymbol{y}))}
$. 
Theorem 1 in  \cite{kawaguchi18analytical}  provides an upper bound on the generalization gap $\Delta_g$
in terms of data distribution in the latent space and properties of the decoder.
\begin{theorem}[\cite{kawaguchi18analytical}]
For any $\ell$, let $(\mathcal{T},f)$be a pair such that $\mathcal{T}:(\mathcal{Z},\mathcal{S})\rightarrow ([0,1]^d,$ $\mathcal{B}([0,1]^d))$ is a measurable function, $f:([0,1]^d,\mathcal{B}([0,1]^d))\rightarrow (\mathbb{R},\mathcal{B}(\mathbb{R}))$ is of bounded variation as $V[f]<\infty$, and
$\ell(\boldsymbol{x}, h(\boldsymbol{y}))=(f\circ \mathcal{T})(\boldsymbol{z}) \forall \boldsymbol{z} \in \mathcal{Z}$, where $\mathcal{B}(A)$ indicates the Borel $\sigma$- algebra on $A$. Then for any dataset pair $(D_n,Z_m)$ and any $\ell(\boldsymbol{x},h_{\mathcal{A}(D_n)}(\boldsymbol{y}))$,
\begin{align*}
\Delta_g={E_{\mu}\ell(\boldsymbol{x},h_{\mathcal{A}(D_n)}(\boldsymbol{y}))}-{E_{Z_m}\ell(\boldsymbol{x},h_{\mathcal{A}(D_n)}(\boldsymbol{y}))}
\leq V[f]\mathcal{D}^*[\mathcal{T}_*\mu, \mathcal{T}(Z_m)]
\end{align*}
where $\mathcal{T}_*\mu$ is pushforward measure of $\mu$ under the map $\mathcal{T}$.
\end{theorem}

For an encoder-decoder setup, $\mathcal{T}$ is the encoder that maps the observation to the latent space and $f$ becomes the composition of loss function and decoder that maps the latent representation to the reconstruction loss. 
Theorem 1 provides two ways to decrease the generalization gap in our problem: by decreasing the variation $V[f]$ or the  discrepancy $\mathcal{D}^*[\mathcal{T}_*\mu, \mathcal{T}(Z_m)]$. Here, we show that stochasticity of the latent space helps decrease the variation $V[f]$. 
The variation of $f$ on $[0,1]^d$ in the sense of Hardy and Krause \cite{hardy}  is defined as: 
$
V[f]=\sum_{k=1}^d\sum_{1\leq j_1<...<j_k\leq d}V^k[f_{j_1...j_k}]
$
where $V^k[f_{j_1...j_k}]$ is defined with following proposition.
\begin{prop}[\cite{kawaguchi18analytical}]
Suppose  that $f_{j_1,..j_k}$ is a function for which $\partial_{1,...k} f_{j_1,..j_k}$ exists on $[0,1]^k$. Then,
$V^k[f_{j_1...j_k}]\leq \underset{\boldsymbol{t}_{j_1},..,\boldsymbol{t}_{j_k}\in [0,1]^k}{sup}|\partial_{1,...k} f_{j_1,..j_k}(\boldsymbol{t}_{j_1},..,\boldsymbol{t}_{j_k})|$.
If $\partial_{1,...k} f_{j_1,..j_k}$ is also continuous on $[0,1]^k$,
$V^k[f_{j_1...j_k}]=\int_{[0,1]^k}|\partial_{1,...k} f_{j_1,..j_k}(\boldsymbol{t}_{j_1},..,\boldsymbol{t}_{j_k})|dt_{j_1}..dt_{j_k}$.
\end{prop}

In our case, 
$f$ is the prediction error $\ell$ as a function of latent representations 
$\boldsymbol{t}$:
\begin{align}
\label{sumloss}
&\ell(\boldsymbol{x}, h(\boldsymbol{y}))=||\boldsymbol{x}-\boldsymbol{g}_{\boldsymbol{\theta}_2}(\boldsymbol{t})||_{F}^2=\sum_i (\boldsymbol{x}_i-g_i(\boldsymbol{t}))^2
=\sum_i\ell_i
\end{align}
where $||\boldsymbol{a}||_{F}$ denotes the Frobenius norm of matrix $\boldsymbol{a}$, and $\boldsymbol{g}_{\boldsymbol{\theta}_2}$ maps the latent space to 
the estimated $\bar{\boldsymbol{x}}$. 
Theorem 1 and Proposition 1 implies that if the  cross partial derivative of the loss with respect to the latent vector at all order is low in all directions throughout the latent space, then the approximated validation loss would be closer to the actual loss over the true unknown distribution of the dataset. Intuitively, we want the loss curve as a function of latent representation to be flat if we want a good generalization. 

\vspace{-.3cm}
\subsubsection{Using stochastic latent space:}
In our formulation, the latent vector is stochastic with the cost function given by eq.(\ref{loss}). 
Using reparameterization $\boldsymbol{\eta}=\boldsymbol{\sigma}_t\odot\boldsymbol{\epsilon}$, the inner expectation of the first term in the loss function $\mathcal{L}_{IB}$ is given by
\begin{align}
\nonumber
T_1&=E_{\boldsymbol{\epsilon}\sim\mathcal{N}(\boldsymbol{0},\boldsymbol{I})}[\sum_i\frac{1}{\boldsymbol{\sigma}_{xi}^2}(\boldsymbol{x}_i-g_i(\boldsymbol{t}+\boldsymbol{\sigma}_t \odot \boldsymbol{\epsilon}))^2]\\
\nonumber
&=\sum_i \frac{1}{\boldsymbol{\sigma}_{x i}^2}(\boldsymbol{x}_i-g_i(\boldsymbol{t}+\boldsymbol{\eta}))^2
=\sum_i \frac{1}{\boldsymbol{\sigma}_{x i}^2} E_{\boldsymbol{\epsilon}}[\ell_i(\boldsymbol{x}_i,\boldsymbol{t}+\boldsymbol{\eta})]
\end{align}
\begin{result}
\begin{align*}
T_1&=\sum_i  \frac{1}{\boldsymbol{\sigma}_{x i}^2} \Big[ \ell_i(\boldsymbol{x}_i,\boldsymbol{t})+\langle \boldsymbol{\sigma}_t \odot E_{\epsilon}[\boldsymbol{\epsilon}],\frac{\partial}{\partial t}\ell_i(\boldsymbol{x}_i,\boldsymbol{t})\rangle\\
&+\frac{1}{2}\langle[\boldsymbol{\sigma}_t\otimes\boldsymbol{\sigma}_t]\odot E_{\epsilon}[\boldsymbol{\epsilon}\otimes\boldsymbol{\epsilon}], \Big[ \frac{\partial ^2}{\partial \boldsymbol{t}_{j_1},\partial \boldsymbol{t}_{j_2}}\ell_i(\boldsymbol{x}_i,\boldsymbol{t}) \Big]\rangle\\
&+..+\frac{1}{k!}\langle[\boldsymbol{\sigma}_t\otimes^k\boldsymbol{\sigma}_t]\odot E_{\epsilon}[\boldsymbol{\epsilon}\otimes^k\boldsymbol{\epsilon}], \Big[ \frac{\partial ^k}{\partial \boldsymbol{t}_{j_1},..,\partial \boldsymbol{t}_{j_k}}\ell_i(\boldsymbol{x}_i,\boldsymbol{t})\Big]\rangle+.. \Big]
\end{align*}
where $[\boldsymbol{\sigma}_t\otimes^k\boldsymbol{\sigma}_t]$ denotes k order tensor product of a vector $\boldsymbol{\sigma}_t$ by itself.
\end{result} 
\begin{proof}
 Using Taylor series expansion for $\ell_i(\boldsymbol{x}_i,\boldsymbol{t}+\boldsymbol{\eta})$,
 \begin{align}
 \nonumber
 E_{\epsilon}[\ell_i(\boldsymbol{x}_i,\boldsymbol{t}+\boldsymbol{\eta})]&={ E_{\epsilon}\Big[\ell_i(\boldsymbol{x}_i,\boldsymbol{t})+\langle\boldsymbol{\eta},\frac{\partial}{\partial \boldsymbol{t}}\ell_i(\boldsymbol{x}_i,\boldsymbol{t})\rangle
 +\frac{1}{2}\langle[\boldsymbol{\eta}\otimes\boldsymbol{\eta}],\Big[ \frac{\partial ^2}{\partial \boldsymbol{t}_{j_1},\partial \boldsymbol{t}_{j_2}}\ell_i(\boldsymbol{x}_i,\boldsymbol{t}) \Big]\rangle}\\
 \label{taylor}
 &+...+\frac{1}{k!}\langle[\boldsymbol{\eta}\otimes^k\boldsymbol{\eta}], \Big[ \frac{\partial ^k}{\partial \boldsymbol{t}_{j_1},..,\partial \boldsymbol{t}_{j_k}}\ell_i(\boldsymbol{x}_i,\boldsymbol{t})\Big]\rangle+..\Big]
 \end{align}
We move expectation operator inside both brackets and take expectation of only the first term in the inner product. Using $\boldsymbol{\eta}=\boldsymbol{\sigma}_t\odot\boldsymbol{\epsilon}$, we get $E_{\epsilon}[\boldsymbol{\eta}\otimes^k\boldsymbol{\eta}]=[\boldsymbol{\sigma}_t\otimes^k\boldsymbol{\sigma}_t]\odot E_{\epsilon}[\boldsymbol{\epsilon}\otimes^k\boldsymbol{\epsilon}]$. Using these in eq.(\ref{taylor}) yields the required result.
\end{proof}
The first term of Result 1, $\ell_i(\boldsymbol{x}_i,\boldsymbol{t})$ (after ignoring $\frac{1}{\boldsymbol{\sigma}_{x i}^2}$), 
would be the only term in the cost function if the latent space were deterministic. The rest of the terms are additional in stochastic training. 
Each of these terms is an inner product of two tensor, the first being  $[\boldsymbol{\sigma}_t\otimes^k\boldsymbol{\sigma}_t]\odot E_{\epsilon}[\boldsymbol{\epsilon}\otimes^k\boldsymbol{\epsilon}]$, and the second being the $k^{th}$ order partial derivative tensor $\Big[ \frac{\partial ^k}{\partial \boldsymbol{t}_{j_1},..,\partial \boldsymbol{t}_{j_k}}\ell_i(\boldsymbol{x}_i,\boldsymbol{t})\Big]$.  We can thus consider the first tensor as providing penalizing weights to different partial derivatives in the second tensor. Since each inner product is added to the cost, we are minimizing them during optimization. This gives two important implications: 

\vspace{-.2cm}
\begin{enumerate}
\item For sufficiently large samples, $E_{\epsilon}[\boldsymbol{\epsilon}\otimes^k\boldsymbol{\epsilon}]$ must be close to central moments of isotropic Gaussian. However, in practice, the number of samples of $\epsilon$ remains constant. As we move to the higher order moment tensors, we can expect that they do not converge to that of the  standard Gaussian. This, luckily, works in our favor. Since we are minimizing $\frac{1}{k!}\langle[\boldsymbol{\sigma}_t\otimes^k\boldsymbol{\sigma}_t]\odot E_{\epsilon}[\boldsymbol{\epsilon}\otimes^k\boldsymbol{\epsilon}], \Big[ \frac{\partial ^k}{\partial \boldsymbol{t}_{j_1},..,\partial \boldsymbol{t}_{j_k}}\ell_i(\boldsymbol{x}_i,\boldsymbol{t})\Big]\rangle$ for each order, the inner product can be vanished for arbitrary $\epsilon$ only by driving partial derivative tensors towards zero. 
Therefore, 
minimizing the sum of all the inner product for arbitrary $\epsilon$ would minimize most of the terms in the partial derivative tensor. From Proposition 1, 
this corresponds to minimizing the variation of function $\ell_i$, and consequently variation of the total error function $\ell$  according to eq.(\ref{sumloss}). Hence, additional terms in the stochastic latent space formulation contributes to decreasing the variation $V[f]$ and consequently the generalization gap.

\item Not all the partial derivatives are equally weighted in the cost function. Due to the presence of weighting tensor $[\boldsymbol{\sigma}_t\otimes^k\boldsymbol{\sigma}_t]$ in the first tensor of inner product, different partial derivative terms are penalized differently according to the value of $\boldsymbol{\sigma}_t$. 
Combination of the KL divergence term in eq.(\ref{loss}) 
with $T_1$ tries to increase standard deviation, $\boldsymbol{\sigma}_t$ towards 1 
whenever it does not significantly increase the cost $T_1$:  higher value of $\boldsymbol{\sigma}_t$ 
penalizes the partial derivatives of a certain direction more heavily, 
making the cost flatter in some directions than other.
\end{enumerate}

\vspace{-.2cm}
Strictly speaking, Proposition 1 requires cross partial derivatives to be small throughout the domain of latent variable, which is not included in the above analysis. 
It however should not significantly affect the observation 
that, compared to deterministic formulation, the stochastic formulation decreases the variation $V[f]$.

\vspace{-.3cm}
\section{Experiments \& Results}
\vspace{-.2cm}
Since it is not possible to obtain real TMP data, the reconstruction network is trained on simulated data pairs of $\boldsymbol{y}$ and $\boldsymbol{x}$. We focus on evaluating three generalization tasks of the network: to learn how to reconstruct under the prior physiological knowledge given in simulation data while generalizing to 1) unseen pathological conditions in $\boldsymbol{x}$, 2) unseen geometrical variations in $\boldsymbol{y}$ that are irrelevant to $\boldsymbol{x}$, and 3) real clinical data. 

\vspace{-.2cm}
\subsection{Generalizing outside the training distribution of TMP}
\subsubsection{Dataset and implementation details:}
\vspace{-.2cm}
We simulated training and test sets using three human-torso geometry models. Spatiotemporal TMP sequences were generated using the Aliev-Panfilov (AP) model \cite{panfilov96}, and projected to the body-surface potential data with 40dB SNR noises. 
Two parameters were varied when simulating the TMP data: the origin of excitation and abnormal tissue properties representing myocardial scar. 
Training data were randomly selected 
with regard to these two parameters. 
Test data were selected 
such that values in these two parameters 
differed from those used in training 
in four levels: 
1) Scar: Low, Exc: Low, 2) Scar: Low, Exc: High, 3) Scar:High, Exc: Low, and 4) Scar: High, Exc: High, where Scar/Exc indicates the parameter being varied and High/Low denotes the 
level of difference (therefore difficulty) from the training data. For example, Scar: Low, Exc: High test ECG data was simulated with region of scar similar to training but origin of excitation very different from that used in training.


For all four models being compared (svs stochastic/deterministic and svs-L stochastic/deterministic), we used ReLU activation functions in both the encoder and decoder, ADAM optimizer \cite{kingma2014adam}, and a learning rate of $10^{-3}$. 
Each neural network was trained on approximately 2500 TMP simulations on each geometry. In addition to the four neural networks, we included a classic TMP inverse reconstruction method (Greensite) designed to incorporate temporal information \cite{greensite98}. On each geometry, approximately 300 cases were tested for each of the four difficulty levels.
We report the average and standard deviation of the results across all three geometry models.

\begin{figure*}[t!]
\centering
\includegraphics[width=\linewidth]{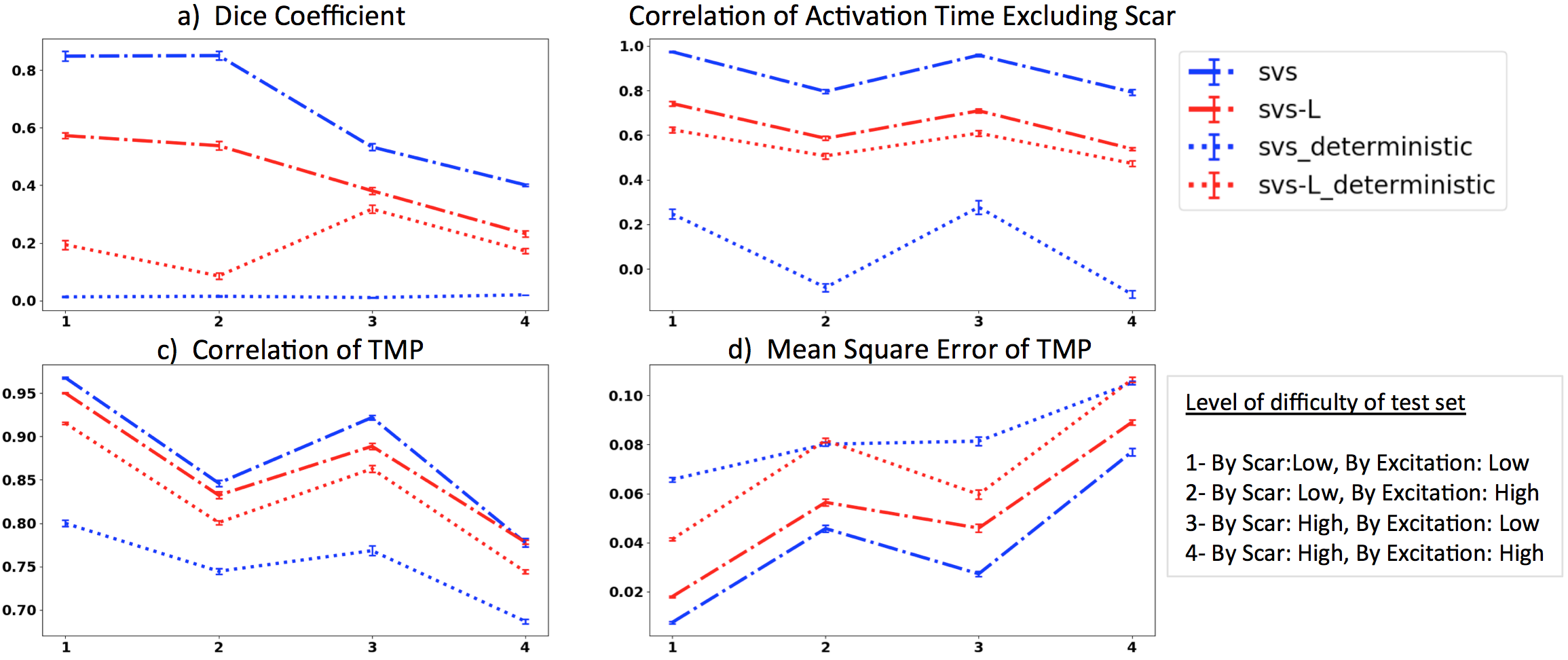}
\vspace{-0.5cm}
\caption{\small{Reconstruction accuracy of different architectures at the presence of test data at different levels of pathological differences from training data.}}
\label{metrics_plot}
\end{figure*}

\vspace{-.4cm}
\subsubsection{Results:}
\begin{table}[tp]
\begin{center}

\caption{\small{Accuracy of different architectures at reconstructing unseen pathological conditions}}
\begin{tabu} to \textwidth {  p{2.5cm} X[c] X[c]  X[c]   X[r]  }
 \hline
  Method \textbackslash  Metric & MSE & TMP Corr. & AT Corr. & Dice Coeff.  \\  
 \hline
 \hline
 \renewcommand{\arraystretch}{1.5}
 \begin{tabular}{@{}c@{}}svs  stochastic\end{tabular} & $\mathbf{0.037} \pm \mathbf{0.021}$ & $\mathbf{0.885} \pm \mathbf{0.061}$ & $\mathbf{0.885} \pm \mathbf{0.072}$& $\mathbf{0.645} \pm \mathbf{0.181}$ \\
 \hline
 \renewcommand{\arraystretch}{1.5}
 \begin{tabular}{c} svs  deterministic\end{tabular}& $0.075 \pm 0.013$ & $0.77 \pm 0.038$ & $0.12 \pm 0.13$& $0.01 \pm 0.006$\\ 
 \hline
 \renewcommand{\arraystretch}{1.5}
 \begin{tabular}{@{}c@{}}svs-L  stochastic\end{tabular}& $0.068 \pm 0.023$ & $0.838 \pm 0.053$ & $0.601 \pm 0.074$& $0.28 \pm 0.154$\\
 \hline
 \renewcommand{\arraystretch}{1.5}
 \begin{tabular}{@{}c@{}}svs-L  deterministic\end{tabular} & $0.067 \pm 0.02$ & $0.84 \pm 0.053$ & $0.57 \pm 0.052$& $0.165 \pm 0.092$ \\ 
 \hline
 \renewcommand{\arraystretch}{1.5}
 \begin{tabular}{@{}c@{}}Greensite\end{tabular} & -- & -- & $0.514 \pm 0.006$& $0.138 \pm 0.005$ \\ 
 \hline
\end{tabu}
\end{center}
\vspace{-1cm}
\end{table}

\begin{figure*}[t!]
\centering
\includegraphics[width=0.8\linewidth]{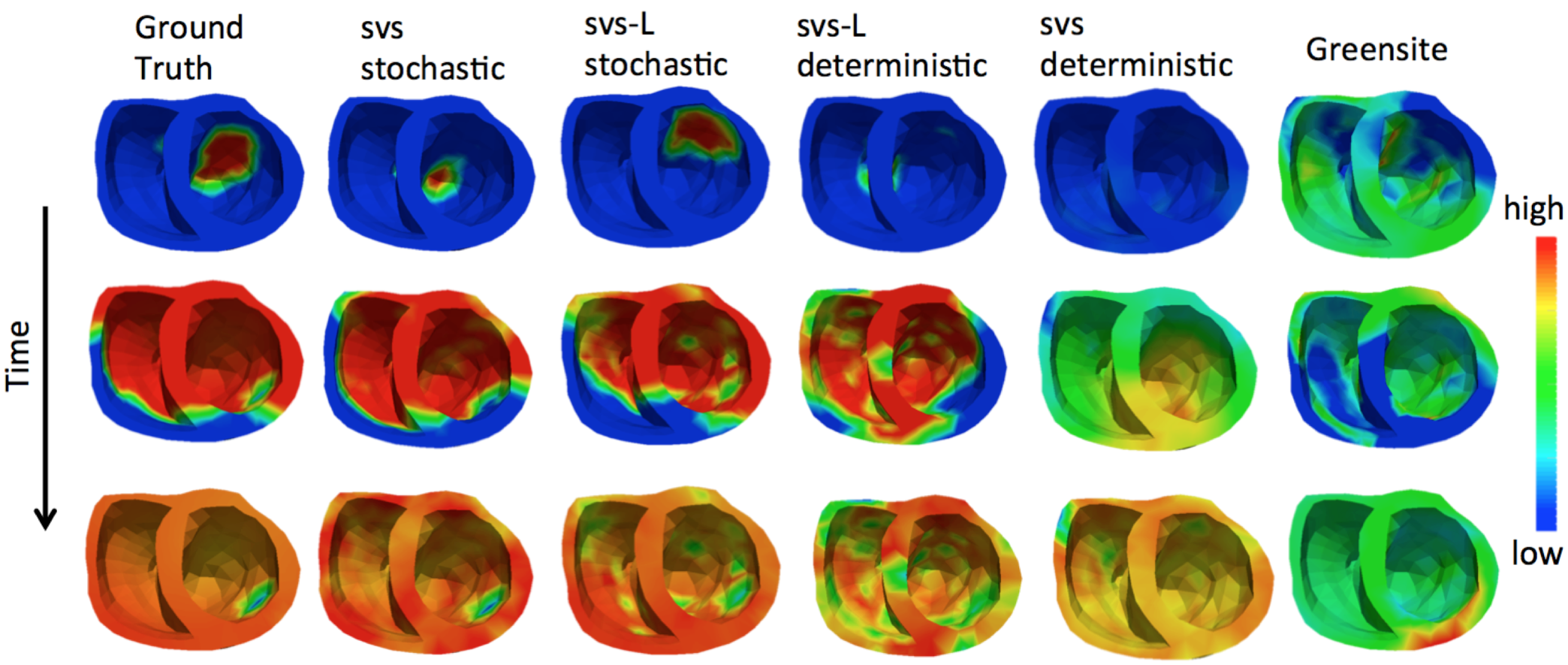}
\caption{\small{Examples of TMP sequences reconstructed by different methods being compared.}}
\label{tmp}
\vspace{-0.2cm}
\end{figure*}

The reconstruction accuracy was measured with four metrics: 1) mean square error (MSE) of the TMP sequence, 2) correlation of the TMP sequence, 3) correlation of TMP-derived activation time (AT),  
and 
4) dice coefficients of the abnormal scar tissue identified from the TMP sequence. 
As summarized in Figure \ref{metrics_plot} and Table 1, 
in all test cases with different levels of pathological differences from the training data, the stochastic version of each architecture was consistently more accurate than its deterministic counterpart. In addition, most of the networks delivered a higher accuracy than the 
classic Greensite method (which does not preserve TMP signal shape and thus its MSE and correlation of TMP was not reported), and the accuracy of the \textit{svs} stochastic architecture was significantly higher than the other architectures.
These observations are reflected in the examples of reconstructed TMP sequences in Fig.~\ref{tmp}. 

\vspace{-.3cm}
\subsection{Generalization to geometrical variations irrelevant to TMP}
\vspace{-.1cm}
\subsubsection{Dataset and implementation details:}
TMP data were simulated as described in the previous section, 
but on a single heart-torso geometry. 
ECG data were simulated from TMP with controlled geometrical variations 
by rotating the heart along Z-axis at different angles (-20 degree to +20 degree at the interval of 1 degree). We trained the network to reconstruct TMP using ECG simulated by i) using five rotation angles from -2 degree to 2 degree, ii) ten rotation angles from -4 degree to +5 degree. We then compared the stochastic and deterministic \textit{svs} networks 
on test ECG generated by the rest of the rotation angles. The network architecture and training details were the same as described in the previous section. 
Test ECG sets at each rotation angle were generated from 250 TMP signals with different tissue properties and origins of excitation and we report the mean and standard deviation of results for each angle.

\begin{figure}[tb!]
\centering
\includegraphics[width=\linewidth]{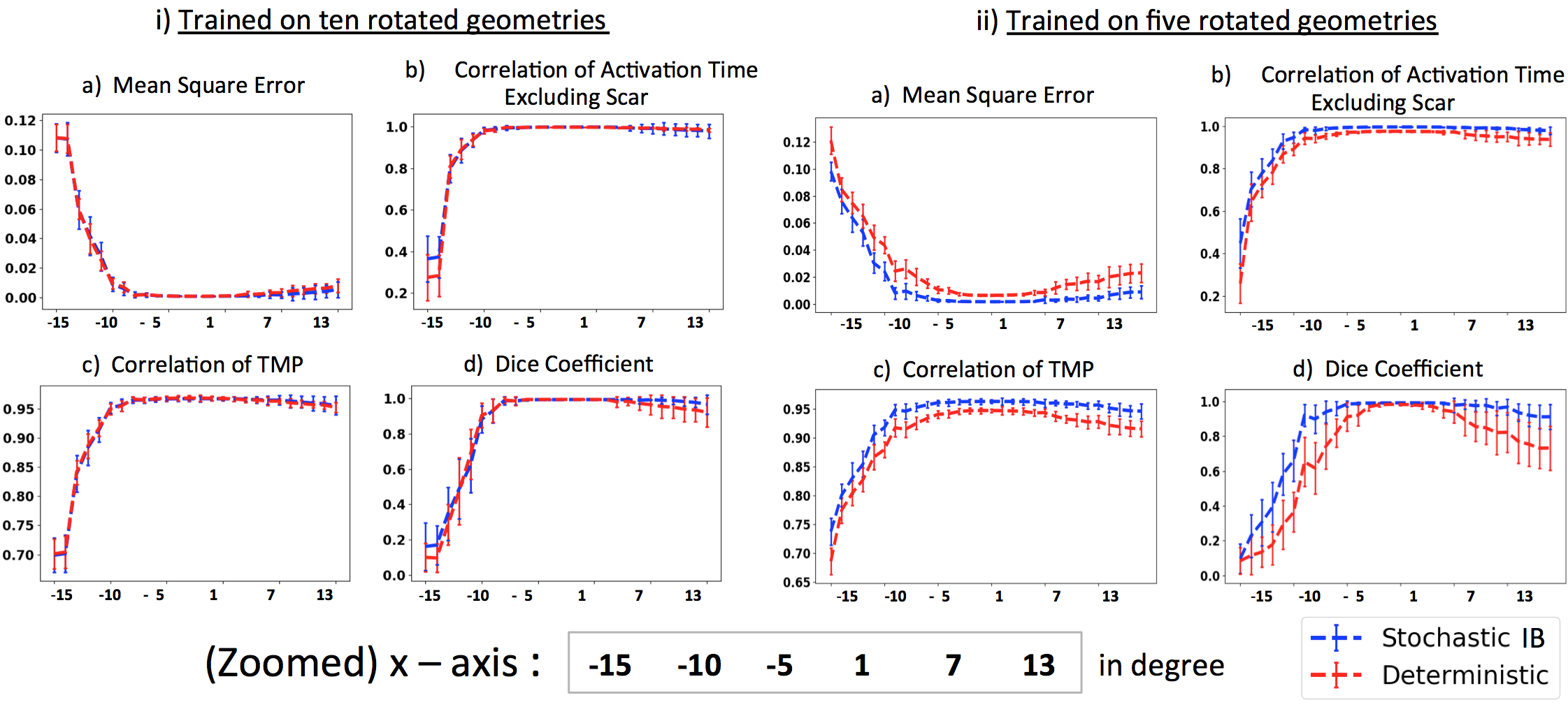}
\vspace{-0.5cm}
\caption{\small{Comparison of TMP reconstruction by stochastic \textit{vs}.\ deterministic networks using training data with a i) high and ii) low amount of variations in geometrical factors irrelevant to TMP. Values along the x axis shows the degree of rotation of the heart relative to the training set, \textit{i.e.,} cases in the center of the x-axis are the closest to the training data. }\vspace{-.3cm}}
\label{low_training}
\end{figure}

\begin{figure}[tb!]
\centering
\includegraphics[width=\linewidth]{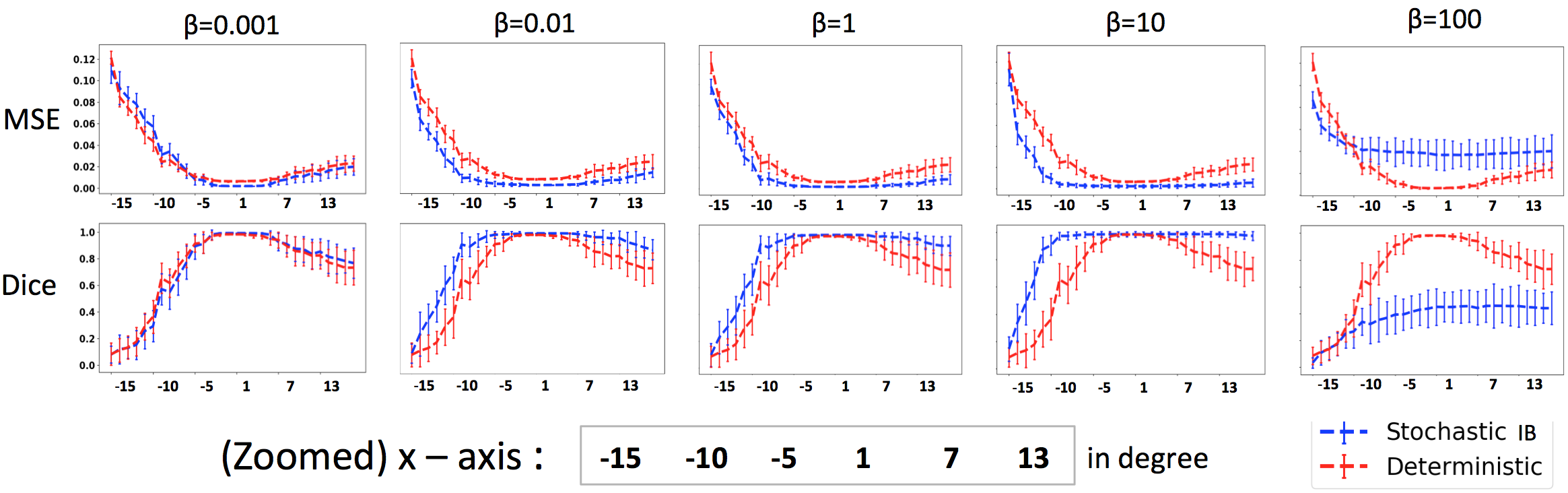}
\vspace{-0.5cm}
\caption{\small{Comparison of stochastic \textit{vs}.\ deterministic architectures at different values of $\beta$. At $\beta=10$, the error stays low and flat for a large range of deviation in angles in stochastic architecture. }\vspace{-.2cm}}
\label{beta_effect}
\end{figure}

\vspace{-.5cm}
\subsubsection{Results:}
As summarized in Fig. \ref{low_training}(ii), 
when trained on a small interval of five rotation values, 
the stochastic information bottleneck consistently 
improves the ability of the network to generalize to geometrical values outside the training distribution. 
This margin of improvement 
also increases as we move further away from the training set, \textit{i.e.} as we go left or right from the centre, and seems to be more pronounced when measuring the dice coefficient of the detected scar.  
When trained on a larger interval of ten rotation values, 
however, 
this performance gap diminishes as shown in Fig. \ref{low_training}(i).  
This suggests that   
the encoder-decoder architecture with compressed latent space can naturally learn to remove variations irrelevant to the network output, 
although the use of stochastic information bottleneck allows the network to generalize from a smaller number of training examples. 

To understand how the parameter $\beta$ in the information bottleneck loss $\mathcal{L}_{IB}$ plays a role in generalization, we repeated the above experiments with different values of $\beta$. As shown in Fig. \ref{beta_effect},  as we increase $\beta$, the generalization ability of the network first increases and then degrades reaching optimum value at $\beta=10$. 

\vspace{-0.3cm}
\subsection{Generalization to real data: a feasibility study}
\vspace{-0.2cm}
\begin{figure*}[tb!]
\centering
\includegraphics[width=\linewidth]{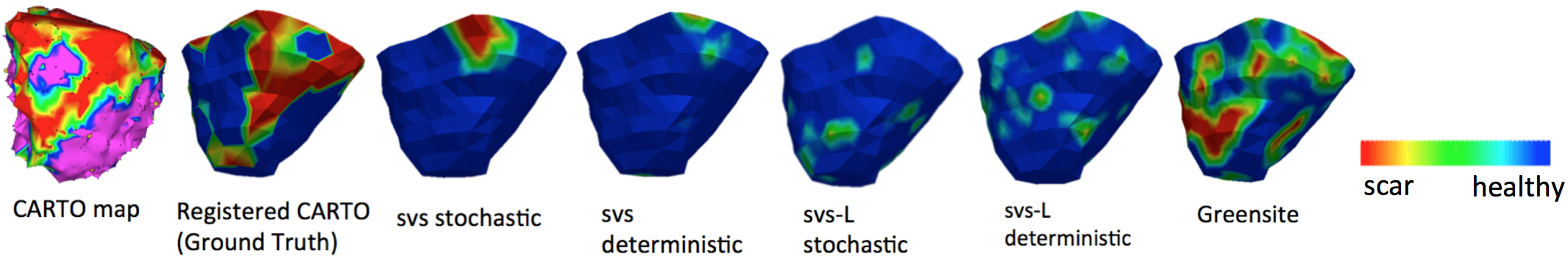}
\vspace{-0.9cm}
\caption{\small{Comparison of scar region identified by different architectures and the Greensite method with reference to \textit{in vivo} voltage maps.}\vspace{-0.4cm}}
\label{realdata}
\end{figure*}
Finally, we tested the presented networks -- trained on simulated data as described earlier --  
on clinical 120-lead ECG data obtained 
from a patient with scar-related ventricular tachycardia. 
From the reconstructed TMP sequence, 
the scar region was delineated based on TMP duration and compared with low-voltage regions from \textit{in-vivo} mapping data. As shown in Fig. \ref{realdata}, 
because the network is directly transferred from the simulated data to real data, 
the reconstruction accuracy is in general lower than that in synthetic cases. However, similar to the observations in synthetic cases, the svs stochastic model 
is able to 
reconstruct 
the region of scar that is the closest to the \textit{in-vivo} data. 
\vspace{-0.3cm}



\section{Conclusion}
\vspace{-0.3cm}
To our knowledge, this is the first work that theoretically investigate the generalization of inverse reconstruction networks through the two different perspectives of stochasticity and information bottleneck, 
supported by carefully designed experiments in 
real-world applications. 
Note that the upper bound 
$\mathcal{L}_{IB} \geq loss_{IB} + D_{KL}(p(\boldsymbol{w})||r(\boldsymbol{w})$. 
Therefore, minimizing $\mathcal{L}_{IB} $ puts an additional constraint on the marginal $p(\boldsymbol{w})$ to be close to a predefined  $r(\boldsymbol{w})$. It is possible that the choice of $r(\boldsymbol{w})$ might also play a role in generalization and will be reserved for future investigations.
Future works will also extend the presented study 
to a wider variety of medical image  reconstruction problems.
\vspace{-.4cm}
\bibliographystyle{splncs04}
\bibliography{bibli1.bib}

\begin{thebibliography}{10}
\providecommand{\url}[1]{\texttt{#1}}
\providecommand{\urlprefix}{URL }
\providecommand{\doi}[1]{https://doi.org/#1}

\bibitem{alemi2017}
Alemi, A., Fischer, I., Dillon, J., Murphy, K.: Deep variational information
  bottleneck. In: ICLR (2017), \url{https://arxiv.org/abs/1612.00410}

\bibitem{panfilov96}
Aliev, R.R., Panfilov, A.V.: A simple two-variable model of cardiac excitation.
  Chaos, Solitons \& Fractals  \textbf{7}(3),  293--301 (1996)

\bibitem{bahdanau14}
Bahdanau, D., Cho, K., Bengio, Y.: Neural machine translation by jointly
  learning to align and translate. arXiv preprint arXiv:1409.0473  (2014)

\bibitem{ghimire2018generative}
Ghimire, S., Dhamala, J., Gyawali, P.K., Sapp, J.L., Horacek, M., Wang, L.:
  Generative modeling and inverse imaging of cardiac transmembrane potential.
  In: International Conference on MICCAI. pp. 508--516. Springer (2018)

\bibitem{greensite98}
Greensite, F., Huiskamp, G.: An improved method for estimating epicardial
  potentials from the body surface. IEEE TBME  \textbf{45}(1),  98--104 (1998)

\bibitem{han2016deep}
Han, Y.S., Yoo, J., Ye, J.C.: Deep residual learning for compressed sensing ct
  reconstruction via persistent homology analysis. arXiv preprint
  arXiv:1611.06391  (2016)

\bibitem{hardy}
Hardy, G.H.: On double fourier series and especially those which represent the
  double zeta-function with real and incommensurable parameters. Quart. J. Math
   \textbf{37}(5) (1906)

\bibitem{kawaguchi18analytical}
Kawaguchi, K., Bengio, Y.: Generalization in machine learning via analytical
  learning theory. arXiv preprint arXiv:1802.07426  (2018)

\bibitem{kingma2014adam}
Kingma, D.P., Ba, J.: Adam: A method for stochastic optimization. ICLR  (2015)

\bibitem{kingma13}
Kingma, D.P., Welling, M.: Auto-encoding variational bayes. ICLR  (2013)

\bibitem{lucas18}
Lucas, A., Iliadis, M., Molina, R., Katsaggelos, A.K.: Using deep neural
  networks for inverse problems in imaging: beyond analytical methods. IEEE
  Signal Processing Magazine  \textbf{35}(1),  20--36 (2018)

\bibitem{luchies2018deep}
Luchies, A.C., Byram, B.C.: Deep neural networks for ultrasound beamforming.
  IEEE transactions on medical imaging  \textbf{37}(9),  2010--2021 (2018)

\bibitem{mao16}
Mao, X., Shen, C., Yang, Y.B.: Image restoration using very deep convolutional
  encoder-decoder networks with symmetric skip connections. In: Advances in
  neural information processing systems. pp. 2802--2810 (2016)

\bibitem{pathak2016context}
Pathak, D., Krahenbuhl, P., Donahue, J., Darrell, T., Efros, A.A.: Context
  encoders: Feature learning by inpainting. In: Proceedings of the IEEE
  Conference on Computer Vision and Pattern Recognition. pp. 2536--2544 (2016)

\bibitem{plonsey1969bioelectric}
Plonsey, R.: Bioelectric phenomena  (1969)

\bibitem{sutskever14}
Sutskever, I., Vinyals, O., Le, Q.V.: Sequence to sequence learning with neural
  networks. In: Advances in neural information processing systems. pp.
  3104--3112 (2014)

\bibitem{tishby2000}
Tishby, N., Pereira, F.C., Bialek, W.: The information bottleneck method. arXiv
  preprint physics/0004057  (2000)

\bibitem{wang10}
Wang, L., Zhang, H., Wong, K.C., Liu, H., Shi, P.:
  Physiological-model-constrained noninvasive reconstruction of volumetric
  myocardial transmembrane potentials. IEEE Transactions on Biomedical
  Engineering  \textbf{57}(2),  296--315 (2010)

\bibitem{zhu18}
Zhu, B., Liu, J.Z., Cauley, S.F., Rosen, B.R., Rosen, M.S.: Image
  reconstruction by domain-transform manifold learning. Nature
  \textbf{555}(7697), ~487 (2018)

\end{thebibliography}


\end{document}